\documentclass[acmtog, nonacm]{acmart}

\renewcommand\footnotetextcopyrightpermission[1]{}
\usepackage{booktabs} 
\usepackage{enumitem}
\usepackage{hyperref}
\usepackage{xcolor}
\usepackage{placeins}
\usepackage{pagesel}

\usepackage[ruled]{algorithm2e} 

\SetAlFnt{\small}
\SetAlCapFnt{\small}
\SetAlCapNameFnt{\small}
\SetAlCapHSkip{0pt}

\makeatletter
\def\@authorsaddresses{}
\makeatother

\begin{document}

\title{MicroZoom: Structure-Preserving Detail Synthesis at Extreme Scale}

\author{Huy Huynh}
\affiliation{%
  \institution{University of Washington}
  \city{Seattle}
  \state{Washington}
  \country{USA}
}

\author{Jingwei Ma}
\affiliation{%
  \institution{University of Washington}
  \city{Seattle}
  \state{Washington}
  \country{USA}
}

\author{Brian Curless}
\affiliation{%
  \institution{University of Washington}
  \city{Seattle}
  \state{Washington}
  \country{USA}
}

\author{Ira Kemelmacher-Schlizerman}
\affiliation{%
  \institution{University of Washington}
  \city{Seattle}
  \state{Washington}
  \country{USA}
}

\author{Steven M. Seitz}
\affiliation{%
  \institution{University of Washington}
  \city{Seattle}
  \state{Washington}
  \country{USA}
}

\begin{abstract}
    We introduce MicroZoom, a generative framework for gigapixel image synthesis at the microscopic scale. Given a standard photograph and a sparse set of consumer-grade microscope close-ups, MicroZoom synthesizes a seamless, gigapixel-resolution image grounded in the material character of the real references, enabling exploratory visualization of microscopic texture across the full spatial extent of an object. Our goal is plausible synthesis, not exact reconstruction. We focus on full-image, reference-based, extreme-scale super-resolution at magnification levels of up to 350$\times$, a setting that introduces two major challenges: (1) recovering texture-specific detail from highly lossy inputs near ambiguous material boundaries, and (2) preserving correct large-scale pattern structure, such as the repeating geometry of a fabric weave, across millions of local predictions. We address these with a two-stage cascaded design, where the first stage recovers global pattern coherence and the second refines local texture detail, supplemented by a segmentation mask to guide synthesis at ambiguous boundaries. We verify our approach on a collection of self-captured everyday objects and demonstrate globally coherent, materially grounded gigapixel imagery. See our \href{https://microzoom-sr.github.io/}{\textcolor{teal}{project page}} for interactive results.
\end{abstract}

\begin{teaserfigure}
  \includegraphics[width=\textwidth]{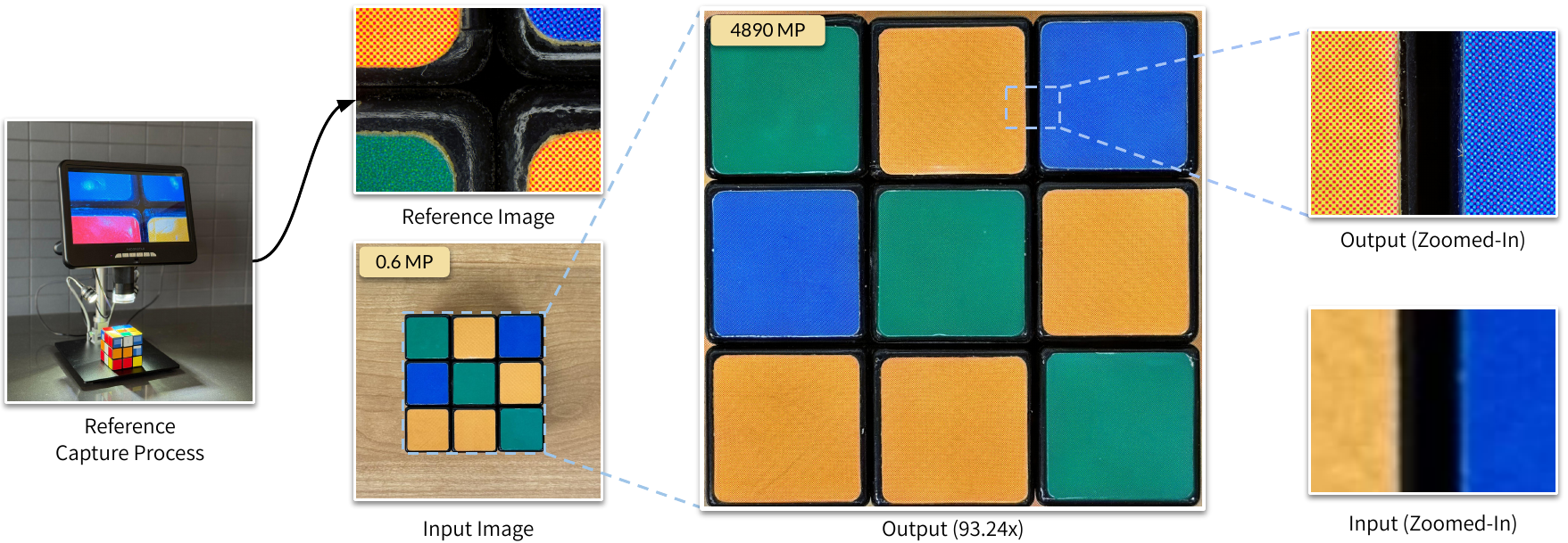}
  \caption{Given a single standard resolution image from a smartphone and a few microscope images, MicroZoom performs extreme-scale super-resolution to synthesize gigapixel imagery. Our method recovers fine-grained textures, such as the patterns on printed paper. This example shows magnification of $93.23\times$, bridging the gap between standard photography and microscopy. See our \href{https://microzoom-sr.github.io/}{\textcolor{teal}{project page}} for interactive results.}
  \label{fig:teaser}
\end{teaserfigure}

\maketitle

\section{Introduction}

The microscopic world is hidden in plain sight: the fine hairs on a leaf, the crystalline structure of salt, and the intricate weave of everyday fabric. These details are present in every object we encounter, yet remain invisible at the scales at which we ordinarily perceive. Bridging the gap between ordinary photographs and microscopically detailed imagery remains an open generative challenge.

Recent generative approaches have made meaningful progress toward this goal. Methods such as Chain-of-Zoom and UltraZoom \cite{wang2024generativepowers,kim2025chainofzoomextremesuperresolutionscale,ma2025ultrazoomgeneratinggigapixelimages} leverage diffusion priors \cite{rombach2022highresolutionimagesynthesislatent, peebles2023scalablediffusionmodelstransformers, saharia2022photorealistictexttoimagediffusionmodels, better_captions_2023} and close-up observations as real-world appearance references. While these methods produce impressive gigapixel imagery from casually captured photographs, they are primarily designed for moderate zoom levels. In this work, we investigate what happens when this paradigm is pushed to substantially larger magnification factors (65--350$\times$).

At extreme scales, two failure modes emerge that prior methods do not adequately address. First, the low-resolution input is highly lossy, making the synthesis of texture-specific detail extremely difficult, with ambiguity further amplified near material boundaries where different textures share indistinguishable low-frequency signals. Second, at a gigapixel scale, each generated region observes only a limited portion of the underlying structure, causing locally plausible predictions to form globally inconsistent structures when merged. Our goal is therefore not exact detail recovery, but the synthesis of globally coherent and visually plausible detail based on real-world references.

To address these challenges, we introduce MicroZoom, a generative framework for extreme-scale detail synthesis from casually captured photographs and high-magnification observations from a commercial digital microscope. Our framework combines two-stage cascaded generation with structure-aware conditioning. The cascade is a deliberate design choice: the first stage synthesizes coherent global structure at a coarser resolution where spatial context is sufficient to establish pattern regularity. This provides the second stage with a structurally sound canvas on which to refine local texture. Our structure-aware conditioning incorporates a segmentation map to preserve material boundaries under extreme magnification.

\vspace{0.3em}
\noindent Our contributions are:
\vspace{0.3em}

\begin{itemize}[topsep=0pt,leftmargin=1.2em]

\item A cascaded diffusion framework for extreme-scale SR that separates global pattern coherence from local detail refinement, enabling structurally regular synthesis at gigapixel resolution.
\item A segmentation-conditioning mechanism that enforces region-consistent texture generation, especially near material boundaries where low-resolution observations become highly ambiguous.
\item A dataset pipeline based on digital microscope focus stacks that produces sharp, extended-depth-of-field references for both training and evaluation under extreme magnification factors.
\end{itemize}

\section{Related Work}

\subsection{Single-Image Super-Resolution}

Single-image super-resolution (SISR) has progressed through successive deep learning paradigms. Dong et al. pioneered end-to-end CNN-based SR with SRCNN \cite{dong2014image}, with residual architectures like EDSR pushing reconstruction accuracy further \cite{lim2017enhanced}. Perceptual quality diverged from reconstruction metrics with SRGAN and ESRGAN, which introduced adversarial and perceptual losses with Residual-in-Residual Dense Blocks to recover photo-realistic textures \cite{ledig2017photo,wang2018esrgan}.

Transformer-based models advanced the field further with SwinIR and HAT, which outperform CNN-based methods across multiple restoration tasks while reducing parameter count \cite{liang2021swinir, chen2025hathybridattentiontransformer}. On the real world degradation front, Real-ESRGAN introduced high-order degradation modeling to handle complex, unknown pipelines \cite{wang2021realesrgan, zhang2021designingpracticaldegradationmodel}, while diffusion-based methods have since pushed perceptual fidelity further: StableSR fine-tunes a time-aware encoder on a frozen Stable Diffusion model \cite{wang2023stablesr}, and DiffBIR decouples degradation removal from detail generation in a two-stage diffusion pipeline \cite{lin2023diffbir}. These methods form the backdrop against which our extreme-scale, reference-based setting is substantially harder: standard SISR assumes modest scale factors and either ignores or hallucinates high-frequency detail, whereas MicroZoom grounds synthesis in real microscope references across 65$\times$-350$\times$ gaps.

\subsection{Reference-Based Super-Resolution}

Reference-based super-resolution (RefSR) synthesizes high-quality detail by transferring textures from an HR reference to an LR input. Freeman et al. proposed example-based SR, retrieving patches from a database \cite{988747}, scaled by Hays et al., using internet-scale scene matching \cite{Sun2012SuperresolutionFI}. Deep RefSR methods split into two paradigms: those transferring high-level style rather than exact textures, and those assuming precise pixel alignment. SRNTT matches and swaps multi-scale features between LR and reference images \cite{zhang2019imagesuperresolutionneuraltexture}, while TTSR leverages a texture transformer for deep feature correspondences \cite{yang2020learningtexturetransformernetwork}. Alignment-focused methods such as C2Matching, DATSR, and DEF robustly warp references onto the input to transfer exact detail \cite{jiang2021robustreferencebasedsuperresolutionc2matching,cao2022referencebasedimagesuperresolutiondeformable,wang2024detailenhancingframeworkreferencebasedimage}, with LMR extending this to multiple references for improved coverage \cite{zhang2023lmrlargescalemultireferencedataset}.

Most relevant to our approach is UltraZoom, which targets mid-range scaling factors, but does not address the structural coherence and domain gap challenges that emerge at microscopic extremes \cite{ma2025ultrazoomgeneratinggigapixelimages}.

 



\subsection{Extreme-Scale Super-Resolution}

Extreme-scale SR targets magnification factors far beyond standard $4\times$ or $8\times$ settings. Generative Powers of Ten and Chain of Zoom (CoZ) generate consistent zooming sequences through cascaded hallucinations, though without real reference grounding \cite{wang2024generativepowers, kim2025chainofzoomextremesuperresolutionscale}. WonderZoom extends CoZ to simultaneous RGB and depth SR but inherits its scale caps and texture-region artifacts \cite{cao2025wonderzoommultiscale3dworld}. Implicit neural representation methods like ContinuousSR model continuous resolution but degrade when evaluated at scales beyond their training range \cite{peng2025pixelgaussianultrafastcontinuous}.


\subsection{Segmentation-Guided Super-Resolution}

Segmentation priors regularize SR by providing spatial structure on content types. SFT-GAN uses Spatial Feature Transform (SFT) layers to modulate features with semantic maps \cite{wang2018recoveringrealistictextureimage, park2019semanticimagesynthesisspatiallyadaptive}, while SegSR, PASD, and SeeSR incorporate segmentation with semantic labels or language descriptions to guide diffusion-based generation \cite{xiao2024semanticsegmentationpriordiffusionbased, yang2024pixelawarestablediffusionrealistic, wu2024seesrsemanticsawarerealworldimage}.

These methods improve category-level spatial coherence but not instance-specific texture fidelity. In MicroZoom, segmentation separates texture regions and prevents cross-region bleeding at material boundaries rather than injecting semantic information.



\section{Method}

\begin{figure*}[!t]
    \centering
    \includegraphics[width=\textwidth]{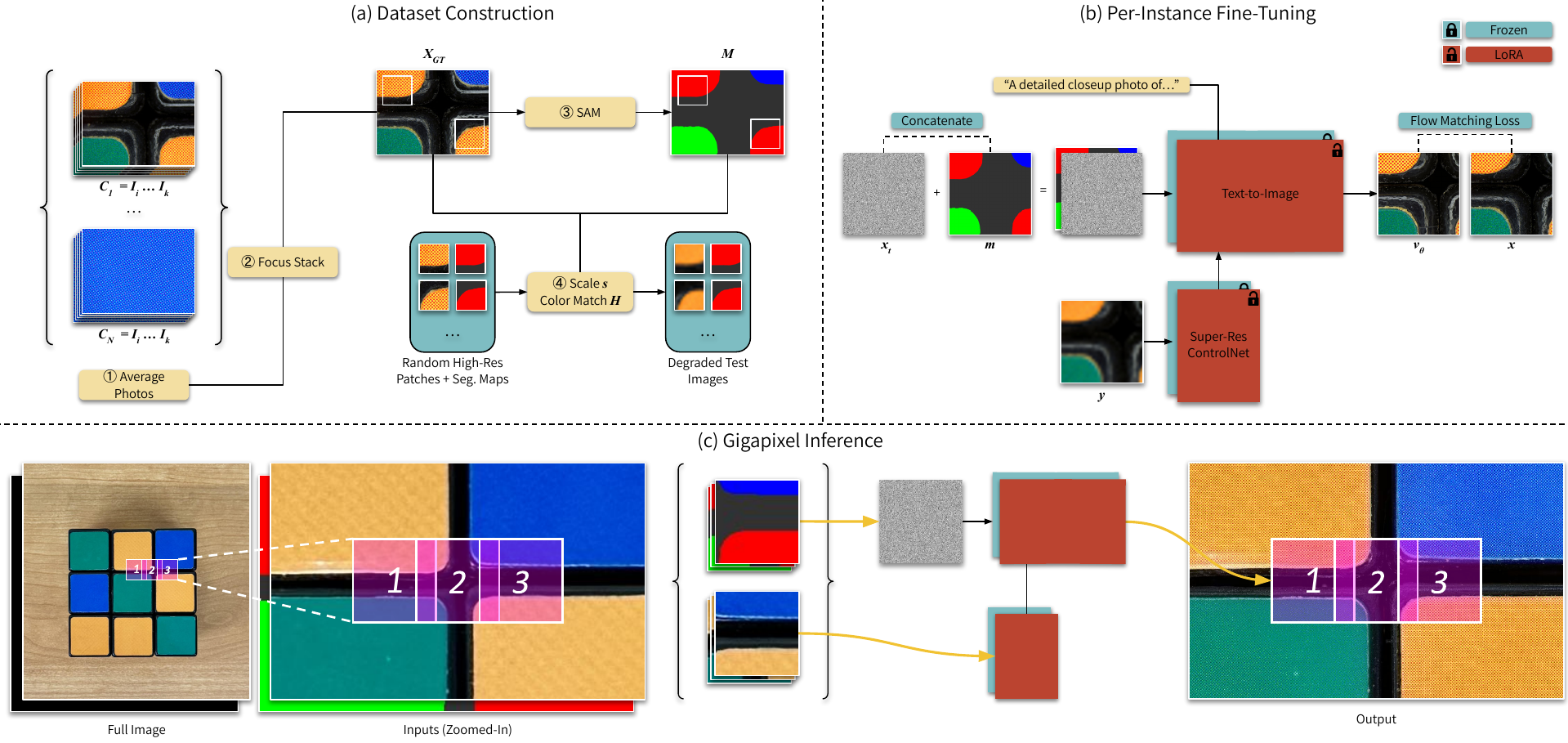}
    \caption{\textbf{Method Overview.} (a) \textit{Dataset Construction}: We curate synthetic training data by capturing a full-view smartphone image and multiple high-magnification digital microscope views. To resolve the shallow depth of field in microscopy, we implement a focus-stacking pipeline that fuses multiple focal planes into a single, sharp ground truth $\mathcal{X}_{GT}$. We then generate aligned training pairs $(\mathcal{X}_{GT}, \mathcal{Y})$ via downsampling and color matching, and compute segmentation masks using SAM to resolve material boundaries. (b) \textit{Per-Instance Fine-Tuning}: We adapt a pre-trained generative model to our super-resolution task \cite{flux2024}. We fine-tune using a parameter-efficient approach, using two main conditioning methods: structure preservation via a ControlNet adapter, and texture synthesis via channel-concatenated segmentation masks. Training follows a conditioning warm-up that prioritizes texture learning before structural alignment (Sec.\ref{method:staged-training}). (c) \textit{Gigapixel Inference}: To generate the final output, we upscale the global input through the cascaded modules sequentially. We utilize MultiDiffusion with a variable-stride sliding window strategy, aggregating local predictions into an artifact-free gigapixel image.}
    \label{fig:pipeline}
\end{figure*}

Given a standard photograph of an object (global view) and a set of microscope close-ups (local views), our goal is to synthesize a giga-pixel resolution image whose large-scale pattern and material character are grounded in the close-up references. This process involves bridging an extreme scale gap (65$\times$ - 350$\times$) while navigating the practical limitations of high-magnification consumer captures, such as shallow depth of field and sensor noise.

Our framework, MicroZoom, addresses this through three stages: dataset construction, boundary-aware per-instance fine-tuning, and gigapixel inference.

\subsection{Dataset Construction}

High-magnification microscopy introduces several unique challenges that are less relevant in standard photography. To collect high-quality ground-truth data for training, we resolve two issues: sensor noise and extremely shallow depth of field. 

\subsubsection{Focus-Stacking and Denoising}

We capture a single full-view image $\mathcal{F}$ using a standard smartphone camera
 and collect $N$ close-up images $\mathcal{C} = {\mathcal{C}_1 \ldots \mathcal{C}_N}$ with local texture details using a digital microscope.

Due to the shallow depth of field at high magnification, a single capture cannot keep textures with height variation (e.g., fabric weaves and wrinkles) uniformly in focus.

To address this, we implement a focus-stacking pipeline. For each microscope view $\mathcal{C}_i$, we capture a $K$-frame focal stack at varying focal distances $\{z_k\}_{k=1}^{K}$. To mitigate sensor noise, each focal plane $\mathcal{I}_{i,z_k}$ is formed by averaging $B$ rapid-burst raw frames $\mathcal{R}$:
\begin{equation}
\mathcal{I}_{i,z_k}=\frac{1}{B}\sum_{j=1}^{B}\mathcal{R}_{i,z_k}^{(j)} .
\end{equation}
We then align the noise-reduced stack using SIFT keypoint matching and fuse the aligned stack via Laplacian Pyramid Fusion, producing a single all-in-focus close-up $\mathcal{C}_i$ \cite{burt1983laplacian, lowe2004distinctive}. We denote the resulting microscope close-ups as $\mathcal{X}_{GT}$ for the rest of the method section.

\subsubsection{Scale Calibration and Synthetic Degradation} 
To construct paired dataset for supervised per-instance super-resolution, we derive training pairs solely from the microscopic ground truth $\mathcal{X}_{GT}$. We include a measurement tool in all captures to estimate the pixel density $\rho$ (pixels/mm). We derive the relative scale factor $s$ as:
\begin{equation}
s = \frac{\rho_{\mathcal{X}_{GT}}}{\rho_{\mathcal{F}}}
\end{equation}
To ensure color consistency between captures, 
we match the color histogram of $\mathcal{X}_{GT}$ to that of the corresponding region in $\mathcal{F}$ in gamma space. We then synthesize the low-resolution input $\mathcal{Y}$ via bicubic downsampling:
\begin{equation}
\mathcal{Y} = \text{Down}_s(\text{ColorMatch}(\mathcal{X}_{GT}, \mathcal{F}))
\end{equation}

\subsubsection{Segmentation Maps}

To resolve ambiguity at material boundaries,
we use the Segment Anything Model (SAM) to generate multi-class segmentation masks $\mathcal{M}$ for each microscopic capture \cite{kirillov2023segment}. These masks are downsampled by the same scale factor $s$ to create $\mathcal{M}_{LR}$, conditioning the model on the semantic layout (e.g. "thread" vs. "button"). Note that segmentation is only applied in multi-material cases where boundary ambiguity is present.

\subsection{Cascaded Per-Instance Fine-Tuning}

At extreme magnification, a single-stage upscaling model faces a fundamental spatial context problem: the low-resolution input contains so little structural information that the model cannot reliably infer the large-scale pattern regularity of the underlying material. The result is output that may be locally plausible, with individual patches that look textured and sharp, but globally incoherent, with repeating structures such as fabric weaves or surface grains that break down across the canvas.

To address this, we adopt a cascaded approach, decomposing the total scale factor $s$ into two sequential stages $s = s_1 \times s_2$. For example, for a $100\times$ target, we train two separate models, $\phi_1$ (coarse stage, e.g., $20\times$) and $\phi_2$ (refinement stage, e.g., $5\times$), and apply them sequentially during inference. The first stage operates on a less extreme upscaling problem, where spatial context is sufficient to recover large-scale pattern structure. The second stage then refines local texture detail on a canvas that is already structurally sound.

This decomposition introduces a known tradeoff: cascading creates a dependency chain, and any structural errors or drift from the first stage are passed into the second. However, this cumulative drift is a far less perceptually harmful failure mode than the globally incoherent patterns produced by single-stage approaches. A slight generative shift in local texture detail is substantially less disruptive to the perceived plausibility of the result. The cascade is therefore a deliberate design choice that trades a bounded, locally-contained error for the structural regularity that makes gigapixel synthesis coherent at scale. See the supplementary material for examples that show global consistency empirically.

Both models are independently trained as described below. For each model, we construct specific training pairs $(x, y)$ where the input $y$ is downsampled by that module's specific factor $s_1$ or $s_2$. We use asymmetric scales because of input dimension constraints. Since we used fixed $1024\times1024$ ground truth crops, downsampling by a factor larger than $5\times$ would result in inputs that fall below the minimum resolution required by the network architecture.

\subsubsection{Training Strategy}

We employ a parameter-efficient fine-tuning approach to adapt a pre-trained generative model to the microscopic domain. We bind a rare identifier token (e.g., "sks") to our specific instance embedded within the text prompt $c_{txt}$ \cite{ruiz2023dreamboothfinetuningtexttoimage, gal2022imageworthwordpersonalizing}. We freeze the pre-trained model weights and inject Low-Rank Adaptation (LoRA) layers into the attention mechanisms \cite{hu2021loralowrankadaptationlarge}.

We define a uniform sampling distribution $\mathcal{D}$ over
 processed inputs. At each training step, we sample $(x, y, m) \sim \mathcal{D}(\mathcal{X}_{GT}, \mathcal{Y}, \mathcal{M}_{LR})$, where $x$ is a random $H \times W$ crop of the high-resolution ground truth patch, $y$ and $m$ are the spatially corresponding low-resolution crop and segmentation mask, respectively.

\subsubsection{Latent Conditioning}

To address the structure-texture trade-off, we introduce two conditioning mechanisms. First, we utilize a ControlNet to guide the global structural alignment conditioned on $y$ \cite{zhang2023addingconditionalcontroltexttoimage}. For material definition, we concatenate the one-hot encoded segmentation mask $m$ directly to the model's noisy latent $x_t$ along the channel dimension. This input is processed by a zero-initialized embedding layer, forcing flow matching to respect material boundaries as a hard constraint.

\subsection{Conditioning Warm-up and Training Objective}
\label{method:staged-training}

We optimize this architecture using a conditioning warm-up, separating structural alignment from texture synthesis. During warm-up, we train for 1000 steps solely on the concatenated segmentation map $m$, with the ControlNet adapter disabled. Then we enable the ControlNet and train for an additional 1000 steps, this time with the full conditioning input $(x, y, m)$. 


The training objective is a Flow Matching loss. Given a data point $x \sim p_{data}$ and noise $x_0 \sim p_{noise}$, we define a probability path $x_t = (1 - t)x_0 + tx$ and a target vector field $u_t(x|x_1) = x - x_0$. The objective minimizes the expected squared error between the model prediction $v_\theta$ and the target vector field:

$$\mathcal{L}_{\text{FM}}(\theta) = \mathbb{E}_{t, x_0, x} \left[ || v_{\theta}(x_t, t, c) - (x - x_0) ||^2 \right]$$

where $c$ represents the aggregate conditioning of text, segmentation, concatenation, and ControlNet features.

\subsection{Gigapixel Inference}

At inference, we aim to upscale the full image $\mathcal{F}$. We first upsample $\mathcal{F}$ to the target gigapixel resolution using bicubic interpolation. To maintain global consistency and avoid grid artifacts, we utilize MultiDiffusion and perform denoising updates on local sliding windows, which are then aggregated back into the global canvas \cite{bartal2023multidiffusionfusingdiffusionpaths, lee2023syncdiffusioncoherentmontagesynchronized}. In addition, since standard sliding window
repeats and amplifies the same boundaries across denoising steps,
we use a variable-stride inference strategy outlined by UltraZoom \cite{ma2025ultrazoomgeneratinggigapixelimages}.

\begin{table}[!t]
    \centering
    \setlength{\tabcolsep}{3.5pt}
    \begin{tabular}{l|cc|cc|cc}
        & \multicolumn{2}{c|}{Syn-LR} & \multicolumn{4}{c}{Ref-LR} \\
        \cline{2-7}
        \rule{0pt}{2.5ex} Method & DISTS $\downarrow$ & LSD $\downarrow$ & \multicolumn{2}{c|}{Win Rate $\uparrow$} & \multicolumn{2}{c}{Top-1 $\uparrow$} \\
        & & & Qual. & Consis. & Qual. & Consis. \\
        \hline
       ContSR  & 0.296 & 1.774 & 0\% & 0\% & 2.05\% & 2.89\% \\
       IPAdapter & 0.324 & 1.813 & 0\% & 0\% & 0.84\% & 1.99\% \\
       CoZ  & 0.297 & 1.386 & 0\% & 0\% & 2.59\% & 6.15 \% \\
       UltraZoom  & 0.232 & \textbf{0.951} & 23.8\% & 28.6\% & 33.99\% & 36.41\% \\
       Ours  & \textbf{0.213} & 1.014 & \textbf{76.2\%} & \textbf{71.4\%} & \textbf{60.52\%} & \textbf{52.56\%} \\
    \end{tabular}
    \caption{\textbf{Quantitative Comparison.} We compare MicroZoom against state-of-the-art super-resolution methods. Note that the automated metrics (DISTS, LSD) are evaluated on the synthetic dataset (Syn-LR) to allow for pixel-aligned ground truth comparisons. In contrast, the user study metrics (Object Win Rate and Top-1 preference) evaluate outputs generated from real smartphone captures (Ref-LR). Best results are in \textbf{bold}. For full synthetic LR visual examples, see the supplemental material.}
    \label{tab:quant_table1}
\end{table}



\FloatBarrier

\section{Results}

\begin{figure*}[t]
    \centering
    \includegraphics[width=\textwidth]{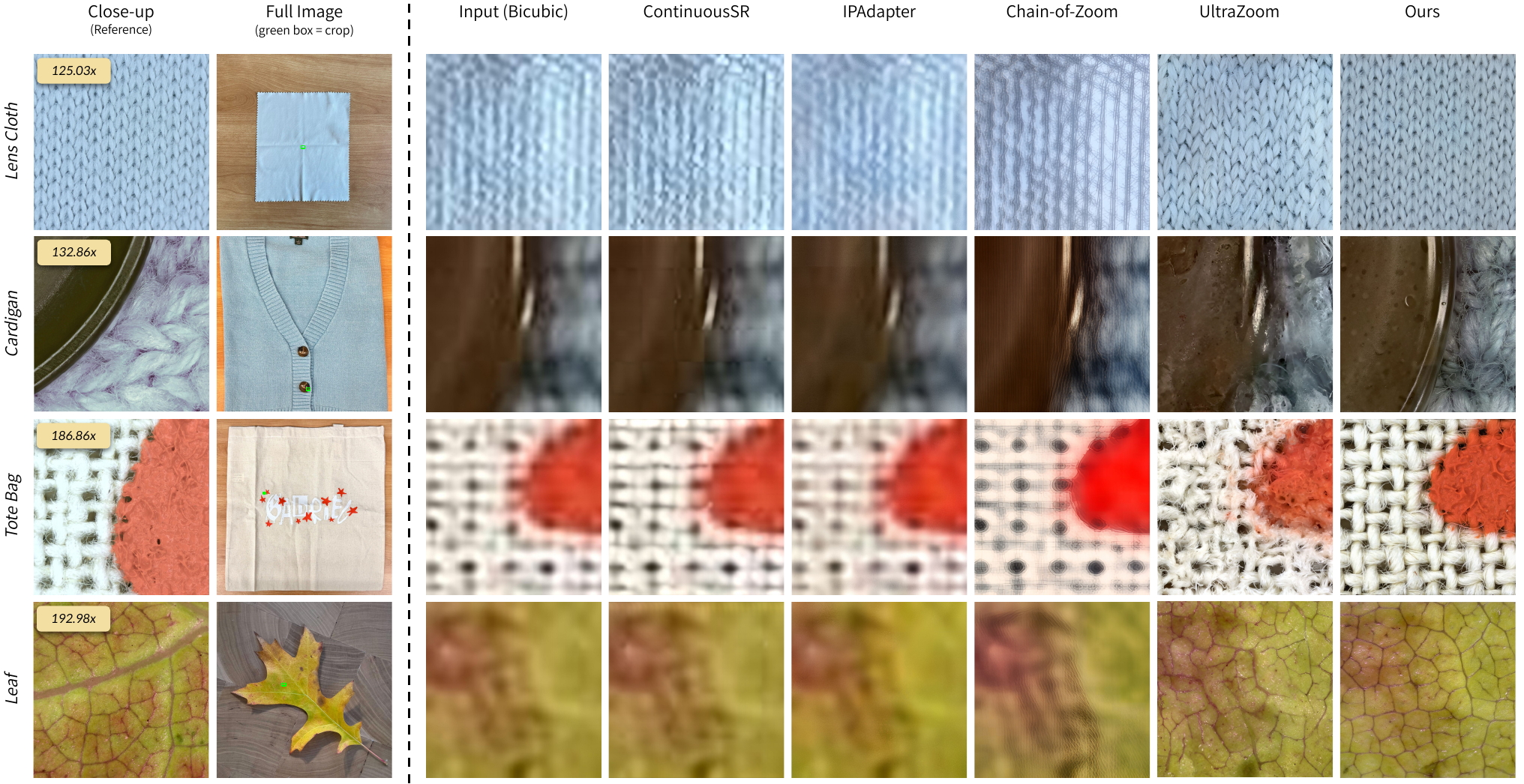}
    \caption{\textbf{Qualitative Comparison.} Rows are ordered from lowest to highest magnification. For each example, we compare a $6000\times6000$ patch across methods. From left to right, we show a microscope capture of the object and the full image with the region of the input patch highlighted. Then, we compare the low-resolution input (upscaled using bicubic interpolation), four baselines, and our final result. These patches are not exactly aligned, but have been selected from similar regions or contain similar materials. Qualitatively, our method achieves the most consistency and visual similarity with the exemplar. For more examples, see Figures \ref{fig:more_multi_texture} and \ref{fig:more_single_texture} or our \href{https://microzoom-sr.github.io/}{\textcolor{teal}{project page}} for full-resolution results.}
    \label{fig:qual}
\end{figure*}

\subsection{Evaluation Data}

We evaluate MicroZoom on a dataset of 21 everyday household items, ranging from organic materials (e.g., leaves, fruits, vegetables) to synthetic structures (e.g., fabrics, plastics). Results for all examples can be found in Figures \ref{fig:qual}, \ref{fig:more_multi_texture}, and \ref{fig:more_single_texture} and in the supplemental material. Each item is captured using the process outlined in Section 3.1, resulting in pairs of full-view images (captured on an iPhone 15 Pro Max) and focus-stacked microscope images (captured on a 48MP digital microscope), with scales ranging from $65\times$ to $350\times$.


\subsection{Computational Cost}

All models are trained on a single A100. We use the Flux.1-dev transformer as our generative backbone \cite{labs2024flux}. We freeze the transformer weights and train only the LoRA adapters (rank $r = 32$). Each training epoch takes approximately $30$ minutes with a batch size of 1. We use the Prodigy optimizer with a learning rate of 1.0. Inference time depends on the input size and scale factor. For example, a $300 \times 300$ pixel image with a scale factor of 132.86$\times$ had an output size of $40000 \times 40000$ and took approximately $20$ hours on a single A100. We note that this cost is not unique to MicroZoom. Any sliding-window diffusion approach operating at gigapixel resolution incurs comparable inference time, as the number of denoising passes scales with the number of local windows.

\subsection{Baseline Methods}

We compare MicroZoom against four state-of-the-art super-resolution methods spanning feed-forward and per-instance paradigms.

ContinuousSR is an arbitrary-scale super-resolution method that models images as a continuous 2D Gaussian Field \cite{peng2025pixelgaussianultrafastcontinuous}. Unlike progressive SR pipelines that require repeated upsampling, ContinuousSR renders outputs directly at the desired scale without architectural modification, making it a natural feed-forward baseline for extreme-scale evaluation.

Chain-of-Zoom (CoZ) is a cascaded framework that progressively upscales images through intermediate resolutions using an autoregressive approach with scale-aware prompts \cite{kim2025chainofzoomextremesuperresolutionscale}. Because CoZ requires a minimum input resolution of $512\times512$, and applies a fixed 4$\times$, we pad inputs to meet size requirements and cascade until exceeding the target resolution, then downsample to the correct final dimensions using area interpolation.

IPAdapter + ControlNet is a feed-forward baseline that conditions a pretrained Flux.1.dev model on reference image via IPAdapter and structural alignment via ControlNet, without any per-instance fine-tuning \cite{ye2023ip-adapter, zhang2023addingconditionalcontroltexttoimage}. This baseline directly tests whether a strong pretrained model with appropriate conditioning can perform reference-guided extreme-scale SR zero-shot, and serves as a point of comparison for the computational cost of per-instance adaptation.


UltraZoom is a reference-based baseline that upsamples a low-resolution input guided by a sparse set of close-up texture references \cite{ma2025ultrazoomgeneratinggigapixelimages}. Like MicroZoom, UltraZoom is a per-instance optimization method; we follow the original protocol and fine-tune it separately for each object.

\begin{figure*}[t!]
    \centering
    \includegraphics[width=\textwidth]{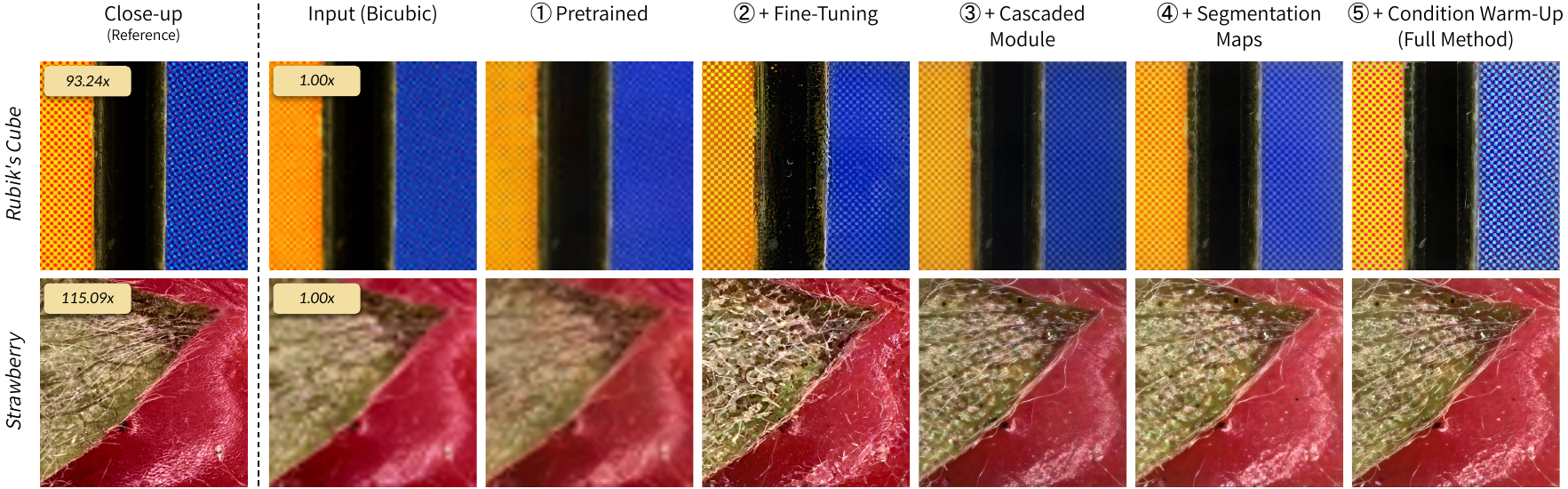}
    \caption{\textbf{Ablations.} We visualize the additive contribution of each module on the final output. Naive super-resolution on a pretrained ControlNet fails to resolve meaningful detail due to the domain gap. Full fine-tuning sharpens textures but suffers from structural hallucinations and material bleeding. Adding segmentation maps resolves these semantic conflicts, guiding boundaries between textures. Finally, our cascaded architecture with conditioning warmup produces the best overall result, maintaining global structural integrity and local texture authenticity.}
    \label{fig:ablation}
\end{figure*}

\subsection{Metrics}

We evaluate using two complementary metrics. DISTS \cite{ding2020iqa} measures perceptual similarity that is tolerant of texture resampling, making it well-suited for generative SR, where outputs are not expected to be pixel-aligned with the ground truth. 
We also report log-spectral distance (LSD), a metric that measures frequency-domain differences between predictions and ground truth. Additional details are provided in the supplementary material.

\subsubsection{Quantitative Comparison}

Table \ref{tab:quant_table1} reports our quantitative metrics (DISTS, LSD) which are calculated exclusively on the synthetic LR split (syn-LR), in which low-resolution inputs are derived by downsampling the ground truth microscope captures, providing pixel-aligned ground truth for quantitative comparison.

MicroZoom achieves the best DISTS (0.213), outperforming UltraZoom (0.232), Chain-of-Zoom (0.297), ContinuousSR (0.296), and IPAdapter+ControlNet (0.324). On LSD, MicroZoom (1.014) is competitive with UltraZoom (0.951), while all other baselines score substantially worse. The feed-forward IPAdapter+ControlNet baseline confirms that zero-shot reference conditioning without per-instance adaptation is insufficient at these magnification factors, scoring last on DISTS (0.324) and LSD (1.813).

\subsubsection{Qualitative Comparison and User Study}

To evaluate perceptual quality beyond automatic metrics, we conducted a user preference study with 79 participants. Unlike the other metrics, the stimuli for this study were generated using our ref-LR pipeline, where standard smartphone photographs serve as the low-resolution input. In each trial, participants were shown the ground-truth microscope patch alongside model-generated patches from all five methods and selected the best output under two criteria: (1) overall visual quality and (2) consistency with the ground truth in terms of material appearance. We included one trial per object, yielding 21 trials and 1,659 votes per criterion.

As seen in Table \ref{tab:quant_table1}, MicroZoom is the most preferred method overall, receiving 60.5\% of quality votes and 52.6\% of similarity votes. Evaluated per object, MicroZoom is the top-ranked method for quality in 16 of 21 objects (76.2\%) and for similarity in 15 of 21 objects (71.4\%). Against ContinuousSR, Chain-of-Zoom, and IPAdapter + ControlNet specifically, MicroZoom wins every single object for both criteria. The closest competition comes from UltraZoom.

These results reflect the qualitative failure modes visible across methods (see Figure \ref{fig:qual}). ContinuousSR, trained on 4--8$\times$ natural image scales, fails to generalize to the narrow field-of-view, texture-heavy close-ups in our dataset, producing outputs that lack sharp definition and material specificity at extreme magnification. 
Chain-of-Zoom's autoregressive cascading amplifies hallucinations at each stage, without real reference grounding, errors compound as the method cascades through 4$\times$ steps, producing results that progressively drifts from the actual object's appearance.

The IPAdapter + ControlNet baseline demonstrates the limitation of feed-forward reference conditioning without per-instance adaptation: while it incorporates both appearance and structural signals, it cannot resolve the extreme domain gap at these magnifications, consistently over-smoothing textures. UltraZoom, which shares our per-instance strategy, captures object identity substantially better than the general baselines but exposes the limitation of single-stage extreme magnification: without the cascaded design and segmentation introduced in MicroZoom, it struggles with multi-material objects, producing regions where textures incorrectly blend or are globally incoherent across the canvas.

\subsection{Ablation Studies}

To validate each component of our pipeline, we conduct both a visual and quantitative ablation, progressively adding components from a pretrained baseline to our full method. Table ~\ref{tab:ablation_quant1} report DISTS and LSD for each quantitative checkpoint. The segmentation map ablation is evaluated on the 11 multi-material objects only, as segmentation conditioning is not applied to single-material cases. The remaining stages are evaluated on all 21 objects.

The most naive baseline applies the pretrained SR model (Flux.1-dev + ControlNet, 4$\times$) without fine-tuning. The significant domain gap between natural images and consumer microscope close-ups at extreme scale results in outputs that are hazy and fail to recover meaningful high-frequency texture, reflected by the worst DISTS (0.330) and LSD (2.055) across all ablation stages.

Per-instance fine-tuning (equivalent to UltraZoom) produces the single largest improvement in the pipeline, with LSD dropping from 2.055 to 0.951 and DISTS improving from 0.330 to 0.232. This reaffirms that instance-specific adaptation is essential at these magnification factors. However, fine-tuning alone performs single-stage extreme upscaling, and without the cascaded design it struggles to maintain structural regularity across multi-material objects, producing texture blending at boundaries.

Adding the cascaded design introduces a slight LSD regression compared to single-scale generation (0.951 $\rightarrow$ 1.021) while improving DISTS (0.232 $\rightarrow$ 0.220), indicating improved perceptual texture and structural similarity.

Adding the segmentation map, evaluated on the 11 multi-material objects, yields a DISTS (0.224) and LSD (1.076) for that subset. MicroZoom's full pipeline achieves the overall best DISTS (0.213) and a competitive LSD (1.014) across the complete 21-object evaluation set. 


\begin{table}[t]
    \centering
    \setlength{\tabcolsep}{15pt}
    \begin{tabular}{l|c|c}
        Stage & DISTS $\downarrow$ & LSD $\downarrow$ \\
        \hline
       Pretrained ControlNet  & 0.330 & 2.055 \\
       + Fine-tuning & 0.232 & \textbf{0.951} \\
       + Cascaded Modules  & 0.220 & 1.021 \\
       + Segmentation Map  & 0.224 & 1.076 \\
       + Cond. Warm Up  & \textbf{0.213} & 1.014 \\
    \end{tabular}
    \caption{\textbf{Ablation Study.} We analyze the quantitative impact of additive modules in our pipeline. Best results are in \textbf{bold}.}
    \label{tab:ablation_quant1}
\end{table}

\FloatBarrier

\subsection{Failure Cases}

\begin{figure}[h!]
    \centering
    \includegraphics[width=\linewidth]{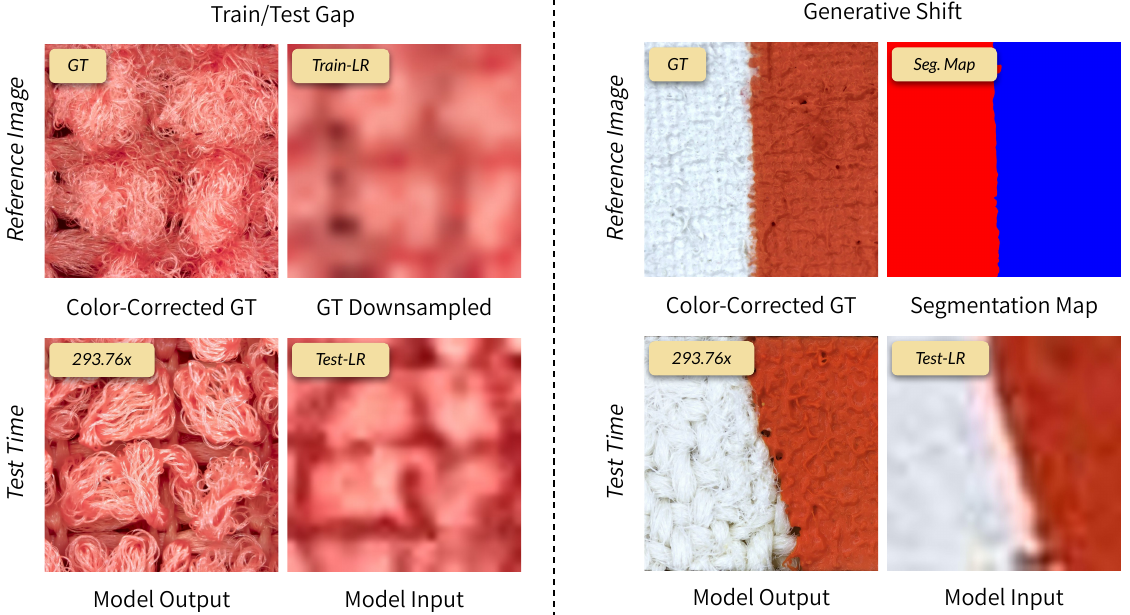}
    \caption{\textbf{Failure Cases.} We visualize two distinct failure cases in our method. The left example showcases the train/test domain gap that exists even after color correction, which leads to inaccurate color shifts. The right example showcases semantic drift, where ambiguous low-frequency information in the input image can generate incorrect textures, even after being labeled.}
    \label{fig:failure}
\end{figure}

 While MicroZoom achieves high-fidelity upscaling, we observe specific limitations and failure cases. See Figure \ref{fig:failure} for a visual example.

 \subsubsection{Train/Test Gap}
 
 Since we fine-tune our model using microscope captures but perform inference with smartphone captures, a domain gap exists between our training and testing data (e.g., different color balance, contrast, and exposure). Despite our color calibration, this discrepancy can lead to color shifts in the final output, particularly in objects with vibrant, saturated hues (e.g., microfiber towel) or in objects with contrasting colors (e.g., strawberry or blue cardigan).

 \subsubsection{Generative Shift}

 Even after fine-tuning, the pipeline can still exhibit generative drift and semantic bleeding. Although we employ segmentation maps to guide texture assignment, we find that the ControlNet acts as a dominant conditioning signal. Consequently, this dense structural guidance can occasionally overpower the segmentation cues, causing textures to incorrectly bleed into one another even when explicitly labeled. 

\section{Discussion and Future Work}

We presented MicroZoom, a generative framework capable of upscaling standard resolution photographs to gigapixel resolution microscopic imagery. Our results show promise 
in addressing hallucination and global-structure degradation that plague extreme-scale super-resolution.
Several constraints point towards future work.

\paragraph{Inference Latency and Computational Cost}

Our inference strategy relies on sliding-window aggregation,
which incurs a substantial computational cost. At these scales, a single image requires 28 denoising steps across tens of thousands of local patches. A single gigapixel output currently requires tens of hours of runtime on an A100 GPU. This latency limits accessibility and applicability. Future research could explore distillation to reduce flow matching inference to a single step, significantly decreasing inference time \cite{song2023consistencymodels, luo2023latentconsistencymodelssynthesizing, sauer2023adversarialdiffusiondistillation}. Alternatively, one could trade accuracy for speed by generating a compact set of repeating texture primitives and retrieving/tiling them, rather than synthesizing every patch independently.

\paragraph{Per-Instance Fine-Tuning} MicroZoom operates as a per-instance generator. 
To minimize drift toward generic textures and preserve instance-specific details, 
we utilize LoRA fine-tuning for each object. This precludes zero-shot generalization to completely unseen materials without retraining. Future works could explore training on a dedicated dataset of microscopic textures to train a general model.

\paragraph{Cascaded Error Accumulation} Our cascaded approach is designed to make each upscaling task easier by recovering macroscale structures before refining microscale details. However, this also introduces a dependency chain. Any errors in boundaries, materials, or hallucinations will be passed into subsequent modules. Future iterations might explore end-to-end models that mitigate the lack of spatial information inherent to extreme-scale super-resolution.


\newpage
\bibliographystyle{ACM-Reference-Format}
\bibliography{sample-bibliography}

@String{Computer = "{IEEE} Computer" }

@misc{bartal2023multidiffusionfusingdiffusionpaths,
      title={MultiDiffusion: Fusing Diffusion Paths for Controlled Image Generation}, 
      author={Omer Bar-Tal and Lior Yariv and Yaron Lipman and Tali Dekel},
      year={2023},
      eprint={2302.08113},
      archivePrefix={arXiv},
      primaryClass={cs.CV},
      url={https://arxiv.org/abs/2302.08113}, 
}

@misc{ma2025ultrazoomgeneratinggigapixelimages,
      title={UltraZoom: Generating Gigapixel Images from Regular Photos}, 
      author={Jingwei Ma and Vivek Jayaram and Brian Curless and Ira Kemelmacher-Shlizerman and Steven M. Seitz},
      year={2025},
      eprint={2506.13756},
      archivePrefix={arXiv},
      primaryClass={cs.CV},
      url={https://arxiv.org/abs/2506.13756}, 
}

@misc{kirillov2023segment,
      title={Segment Anything}, 
      author={Alexander Kirillov and Eric Mintun and Nikhila Ravi and Hanzi Mao and Chloe Rolland and Laura Gustafson and Tete Xiao and Spencer Whitehead and Alexander C. Berg and Wan-Yen Lo and Piotr Dollár and Ross Girshick},
      year={2023},
      eprint={2304.02643},
      archivePrefix={arXiv},
      primaryClass={cs.CV},
      url={https://arxiv.org/abs/2304.02643}, 
}

@misc{ruiz2023dreamboothfinetuningtexttoimage,
      title={DreamBooth: Fine Tuning Text-to-Image Diffusion Models for Subject-Driven Generation}, 
      author={Nataniel Ruiz and Yuanzhen Li and Varun Jampani and Yael Pritch and Michael Rubinstein and Kfir Aberman},
      year={2023},
      eprint={2208.12242},
      archivePrefix={arXiv},
      primaryClass={cs.CV},
      url={https://arxiv.org/abs/2208.12242}, 
}

@misc{hu2021loralowrankadaptationlarge,
      title={LoRA: Low-Rank Adaptation of Large Language Models}, 
      author={Edward J. Hu and Yelong Shen and Phillip Wallis and Zeyuan Allen-Zhu and Yuanzhi Li and Shean Wang and Lu Wang and Weizhu Chen},
      year={2021},
      eprint={2106.09685},
      archivePrefix={arXiv},
      primaryClass={cs.CL},
      url={https://arxiv.org/abs/2106.09685}, 
}

@misc{zhang2023addingconditionalcontroltexttoimage,
      title={Adding Conditional Control to Text-to-Image Diffusion Models}, 
      author={Lvmin Zhang and Anyi Rao and Maneesh Agrawala},
      year={2023},
      eprint={2302.05543},
      archivePrefix={arXiv},
      primaryClass={cs.CV},
      url={https://arxiv.org/abs/2302.05543}, 
}

@misc{labs2024flux,
  title={{FLUX.1}: A family of flow matching models},
  author={{Black Forest Labs}},
  year={2024},
  howpublished={\url{https://blackforestlabs.ai}}
}

@misc{kim2025chainofzoomextremesuperresolutionscale,
      title={Chain-of-Zoom: Extreme Super-Resolution via Scale Autoregression and Preference Alignment}, 
      author={Bryan Sangwoo Kim and Jeongsol Kim and Jong Chul Ye},
      year={2025},
      eprint={2505.18600},
      archivePrefix={arXiv},
      primaryClass={cs.CV},
      url={https://arxiv.org/abs/2505.18600}, 
}

@misc{peng2025pixelgaussianultrafastcontinuous,
      title={Pixel to Gaussian: Ultra-Fast Continuous Super-Resolution with 2D Gaussian Modeling}, 
      author={Long Peng and Anran Wu and Wenbo Li and Peizhe Xia and Xueyuan Dai and Xinjie Zhang and Xin Di and Haoze Sun and Renjing Pei and Yang Wang and Yang Cao and Zheng-Jun Zha},
      year={2025},
      eprint={2503.06617},
      archivePrefix={arXiv},
      primaryClass={cs.CV},
      url={https://arxiv.org/abs/2503.06617}, 
}

@ARTICLE{988747,
  author={Freeman, W.T. and Jones, T.R. and Pasztor, E.C.},
  journal={IEEE Computer Graphics and Applications}, 
  title={Example-based super-resolution}, 
  year={2002},
  volume={22},
  number={2},
  pages={56-65},
  doi={10.1109/38.988747}}

@article{Sun2012SuperresolutionFI,
  title={Super-resolution from internet-scale scene matching},
  author={Libin Sun and James Hays},
  journal={2012 IEEE International Conference on Computational Photography (ICCP)},
  year={2012},
  pages={1-12},
  url={https://api.semanticscholar.org/CorpusID:2311801}
}

@misc{zhang2019imagesuperresolutionneuraltexture,
      title={Image Super-Resolution by Neural Texture Transfer}, 
      author={Zhifei Zhang and Zhaowen Wang and Zhe Lin and Hairong Qi},
      year={2019},
      eprint={1903.00834},
      archivePrefix={arXiv},
      primaryClass={cs.CV},
      url={https://arxiv.org/abs/1903.00834}, 
}

@misc{yang2020learningtexturetransformernetwork,
      title={Learning Texture Transformer Network for Image Super-Resolution}, 
      author={Fuzhi Yang and Huan Yang and Jianlong Fu and Hongtao Lu and Baining Guo},
      year={2020},
      eprint={2006.04139},
      archivePrefix={arXiv},
      primaryClass={cs.CV},
      url={https://arxiv.org/abs/2006.04139}, 
}

@misc{jiang2021robustreferencebasedsuperresolutionc2matching,
      title={Robust Reference-based Super-Resolution via C2-Matching}, 
      author={Yuming Jiang and Kelvin C. K. Chan and Xintao Wang and Chen Change Loy and Ziwei Liu},
      year={2021},
      eprint={2106.01863},
      archivePrefix={arXiv},
      primaryClass={cs.CV},
      url={https://arxiv.org/abs/2106.01863}, 
}

@misc{zhang2023lmrlargescalemultireferencedataset,
      title={LMR: A Large-Scale Multi-Reference Dataset for Reference-based Super-Resolution}, 
      author={Lin Zhang and Xin Li and Dongliang He and Errui Ding and Zhaoxiang Zhang},
      year={2023},
      eprint={2303.04970},
      archivePrefix={arXiv},
      primaryClass={cs.CV},
      url={https://arxiv.org/abs/2303.04970}, 
}

@misc{cao2022referencebasedimagesuperresolutiondeformable,
      title={Reference-based Image Super-Resolution with Deformable Attention Transformer}, 
      author={Jiezhang Cao and Jingyun Liang and Kai Zhang and Yawei Li and Yulun Zhang and Wenguan Wang and Luc Van Gool},
      year={2022},
      eprint={2207.11938},
      archivePrefix={arXiv},
      primaryClass={cs.CV},
      url={https://arxiv.org/abs/2207.11938}, 
}

@misc{wang2024generativepowers,
      title={Generative Powers of Ten}, 
      author={Xiaojuan Wang and Janne Kontkanen and Brian Curless and Steve Seitz and Ira Kemelmacher and Ben Mildenhall and Pratul Srinivasan and Dor Verbin and Aleksander Holynski},
      year={2024},
      eprint={2312.02149},
      archivePrefix={arXiv},
      primaryClass={cs.CV},
      url={https://arxiv.org/abs/2312.02149}, 
}

@misc{cao2025wonderzoommultiscale3dworld,
      title={WonderZoom: Multi-Scale 3D World Generation}, 
      author={Jin Cao and Hong-Xing Yu and Jiajun Wu},
      year={2025},
      eprint={2512.09164},
      archivePrefix={arXiv},
      primaryClass={cs.CV},
      url={https://arxiv.org/abs/2512.09164}, 
}

@misc{wang2018recoveringrealistictextureimage,
      title={Recovering Realistic Texture in Image Super-resolution by Deep Spatial Feature Transform}, 
      author={Xintao Wang and Ke Yu and Chao Dong and Chen Change Loy},
      year={2018},
      eprint={1804.02815},
      archivePrefix={arXiv},
      primaryClass={cs.CV},
      url={https://arxiv.org/abs/1804.02815}, 
}

@misc{xiao2024semanticsegmentationpriordiffusionbased,
      title={Semantic Segmentation Prior for Diffusion-Based Real-World Super-Resolution}, 
      author={Jiahua Xiao and Jiawei Zhang and Dongqing Zou and Xiaodan Zhang and Jimmy Ren and Xing Wei},
      year={2024},
      eprint={2412.02960},
      archivePrefix={arXiv},
      primaryClass={cs.CV},
      url={https://arxiv.org/abs/2412.02960}, 
}

@misc{yang2024pixelawarestablediffusionrealistic,
      title={Pixel-Aware Stable Diffusion for Realistic Image Super-resolution and Personalized Stylization}, 
      author={Tao Yang and Rongyuan Wu and Peiran Ren and Xuansong Xie and Lei Zhang},
      year={2024},
      eprint={2308.14469},
      archivePrefix={arXiv},
      primaryClass={cs.CV},
      url={https://arxiv.org/abs/2308.14469}, 
}

@misc{wu2024seesrsemanticsawarerealworldimage,
      title={SeeSR: Towards Semantics-Aware Real-World Image Super-Resolution}, 
      author={Rongyuan Wu and Tao Yang and Lingchen Sun and Zhengqiang Zhang and Shuai Li and Lei Zhang},
      year={2024},
      eprint={2311.16518},
      archivePrefix={arXiv},
      primaryClass={cs.CV},
      url={https://arxiv.org/abs/2311.16518}, 
}

@misc{wang2024detailenhancingframeworkreferencebasedimage,
      title={Detail-Enhancing Framework for Reference-Based Image Super-Resolution}, 
      author={Zihan Wang and Ziliang Xiong and Hongying Tang and Xiaobing Yuan},
      year={2024},
      eprint={2405.00431},
      archivePrefix={arXiv},
      primaryClass={cs.CV},
      url={https://arxiv.org/abs/2405.00431}, 
}

@article{ye2023ip-adapter,
  title={IP-Adapter: Text Compatible Image Prompt Adapter for Text-to-Image Diffusion Models},
  author={Ye, Hu and Zhang, Jun and Liu, Sibo and Han, Xiao and Yang, Wei},
  booktitle={arXiv preprint arxiv:2308.06721},
  year={2023}
}

@article{ding2020iqa,
  title={Image Quality Assessment: Unifying Structure and Texture Similarity},
  author={Ding, Keyan and Ma, Kede and Wang, Shiqi and Simoncelli, Eero P.},
  journal = {CoRR},
  volume = {abs/2004.07728},
  year={2020},
  url = {https://arxiv.org/abs/2004.07728}
}

@article{lowe2004distinctive,
  title={Distinctive image features from scale-invariant keypoints},
  author={Lowe, David G},
  journal={International Journal of Computer Vision},
  volume={60},
  number={2},
  pages={91--110},
  year={2004}
}

@article{burt1983laplacian,
  title={The Laplacian pyramid as a compact image code},
  author={Burt, Peter J and Adelson, Edward H},
  journal={IEEE Transactions on Communications},
  volume={31},
  number={4},
  pages={532--540},
  year={1983}
}

@article{dong2014image,
  title={Image super-resolution using deep convolutional networks},
  author={Dong, Chao and Loy, Chen Change and He, Kaiming and Tang, Xiaoou},
  journal={arXiv preprint arXiv:1501.00092},
  year={2014}
}

@inproceedings{lim2017enhanced,
  title={Enhanced deep residual networks for single image super-resolution},
  author={Lim, Bee and Son, Sanghyun and Kim, Heewon and Nah, Seungjun and Lee, Kyoung Mu},
  booktitle={CVPR Workshops},
  year={2017}
}

@inproceedings{ledig2017photo,
  title={Photo-realistic single image super-resolution using a generative adversarial network},
  author={Ledig, Christian and Theis, Lucas and Husz{\'a}r, Ferenc and others},
  booktitle={CVPR},
  year={2017}
}

@inproceedings{wang2018esrgan,
  title={{ESRGAN}: Enhanced super-resolution generative adversarial networks},
  author={Wang, Xintao and Yu, Ke and Wu, Shixiang and others},
  booktitle={ECCV Workshops},
  year={2018}
}

@inproceedings{liang2021swinir,
  title={{SwinIR}: Image restoration using swin transformer},
  author={Liang, Jingyun and Cao, Jiezhang and Sun, Guolei and Zhang, Kai and Van Gool, Luc and Timofte, Radu},
  booktitle={ICCV Workshops},
  year={2021}
}

@inproceedings{wang2021realesrgan,
  title={Real-{ESRGAN}: Training real-world blind super-resolution with pure synthetic data},
  author={Wang, Xintao and Xie, Liangbin and Dong, Chao and Shan, Ying},
  booktitle={ICCV Workshops},
  year={2021}
}

@article{wang2023stablesr,
  title={Exploiting diffusion prior for real-world image super-resolution},
  author={Wang, Jianyi and Yue, Zongsheng and Zhou, Shangchen and Chan, Kelvin CK and Loy, Chen Change},
  journal={arXiv preprint arXiv:2305.07015},
  year={2023}
}

@article{lin2023diffbir,
  title={{DiffBIR}: Towards blind image restoration with generative diffusion prior},
  author={Lin, Xinqi and He, Jingwen and Chen, Ziyan and others},
  journal={arXiv preprint arXiv:2308.15070},
  year={2023}
}

@misc{flux2024,
    author={Black Forest Labs},
    title={FLUX},
    year={2024},
    howpublished={\url{https://github.com/black-forest-labs/flux}},
}

@misc{rombach2022highresolutionimagesynthesislatent,
      title={High-Resolution Image Synthesis with Latent Diffusion Models}, 
      author={Robin Rombach and Andreas Blattmann and Dominik Lorenz and Patrick Esser and Björn Ommer},
      year={2022},
      eprint={2112.10752},
      archivePrefix={arXiv},
      primaryClass={cs.CV},
      url={https://arxiv.org/abs/2112.10752}, 
}

@misc{peebles2023scalablediffusionmodelstransformers,
      title={Scalable Diffusion Models with Transformers}, 
      author={William Peebles and Saining Xie},
      year={2023},
      eprint={2212.09748},
      archivePrefix={arXiv},
      primaryClass={cs.CV},
      url={https://arxiv.org/abs/2212.09748}, 
}

@misc{saharia2022photorealistictexttoimagediffusionmodels,
      title={Photorealistic Text-to-Image Diffusion Models with Deep Language Understanding}, 
      author={Chitwan Saharia and William Chan and Saurabh Saxena and Lala Li and Jay Whang and Emily Denton and Seyed Kamyar Seyed Ghasemipour and Burcu Karagol Ayan and S. Sara Mahdavi and Rapha Gontijo Lopes and Tim Salimans and Jonathan Ho and David J Fleet and Mohammad Norouzi},
      year={2022},
      eprint={2205.11487},
      archivePrefix={arXiv},
      primaryClass={cs.CV},
      url={https://arxiv.org/abs/2205.11487}, 
}

@article{better_captions_2023,
  title={Improving Image Generation with Better Captions},
  author={Ramesh, Aditya and Dhariwal, Prafulla and Nichol, Alex and Chu, Casey and Chen, Mark},
  journal={arXiv preprint arXiv:2309.16609},
  year={2023}
}

@misc{chen2025hathybridattentiontransformer,
      title={HAT: Hybrid Attention Transformer for Image Restoration}, 
      author={Xiangyu Chen and Xintao Wang and Wenlong Zhang and Xiangtao Kong and Yu Qiao and Jiantao Zhou and Chao Dong},
      year={2025},
      eprint={2309.05239},
      archivePrefix={arXiv},
      primaryClass={cs.CV},
      url={https://arxiv.org/abs/2309.05239}, 
}

@misc{zhang2021designingpracticaldegradationmodel,
      title={Designing a Practical Degradation Model for Deep Blind Image Super-Resolution}, 
      author={Kai Zhang and Jingyun Liang and Luc Van Gool and Radu Timofte},
      year={2021},
      eprint={2103.14006},
      archivePrefix={arXiv},
      primaryClass={eess.IV},
      url={https://arxiv.org/abs/2103.14006}, 
}

@misc{park2019semanticimagesynthesisspatiallyadaptive,
      title={Semantic Image Synthesis with Spatially-Adaptive Normalization}, 
      author={Taesung Park and Ming-Yu Liu and Ting-Chun Wang and Jun-Yan Zhu},
      year={2019},
      eprint={1903.07291},
      archivePrefix={arXiv},
      primaryClass={cs.CV},
      url={https://arxiv.org/abs/1903.07291}, 
}

@misc{gal2022imageworthwordpersonalizing,
      title={An Image is Worth One Word: Personalizing Text-to-Image Generation using Textual Inversion}, 
      author={Rinon Gal and Yuval Alaluf and Yuval Atzmon and Or Patashnik and Amit H. Bermano and Gal Chechik and Daniel Cohen-Or},
      year={2022},
      eprint={2208.01618},
      archivePrefix={arXiv},
      primaryClass={cs.CV},
      url={https://arxiv.org/abs/2208.01618}, 
}

@misc{lee2023syncdiffusioncoherentmontagesynchronized,
      title={SyncDiffusion: Coherent Montage via Synchronized Joint Diffusions}, 
      author={Yuseung Lee and Kunho Kim and Hyunjin Kim and Minhyuk Sung},
      year={2023},
      eprint={2306.05178},
      archivePrefix={arXiv},
      primaryClass={cs.CV},
      url={https://arxiv.org/abs/2306.05178}, 
}

@misc{song2023consistencymodels,
      title={Consistency Models}, 
      author={Yang Song and Prafulla Dhariwal and Mark Chen and Ilya Sutskever},
      year={2023},
      eprint={2303.01469},
      archivePrefix={arXiv},
      primaryClass={cs.LG},
      url={https://arxiv.org/abs/2303.01469}, 
}

@misc{luo2023latentconsistencymodelssynthesizing,
      title={Latent Consistency Models: Synthesizing High-Resolution Images with Few-Step Inference}, 
      author={Simian Luo and Yiqin Tan and Longbo Huang and Jian Li and Hang Zhao},
      year={2023},
      eprint={2310.04378},
      archivePrefix={arXiv},
      primaryClass={cs.CV},
      url={https://arxiv.org/abs/2310.04378}, 
}

@misc{sauer2023adversarialdiffusiondistillation,
      title={Adversarial Diffusion Distillation}, 
      author={Axel Sauer and Dominik Lorenz and Andreas Blattmann and Robin Rombach},
      year={2023},
      eprint={2311.17042},
      archivePrefix={arXiv},
      primaryClass={cs.CV},
      url={https://arxiv.org/abs/2311.17042}, 
}

\newpage

\begin{figure*}[t]
    \centering
    \includegraphics[width=\textwidth]{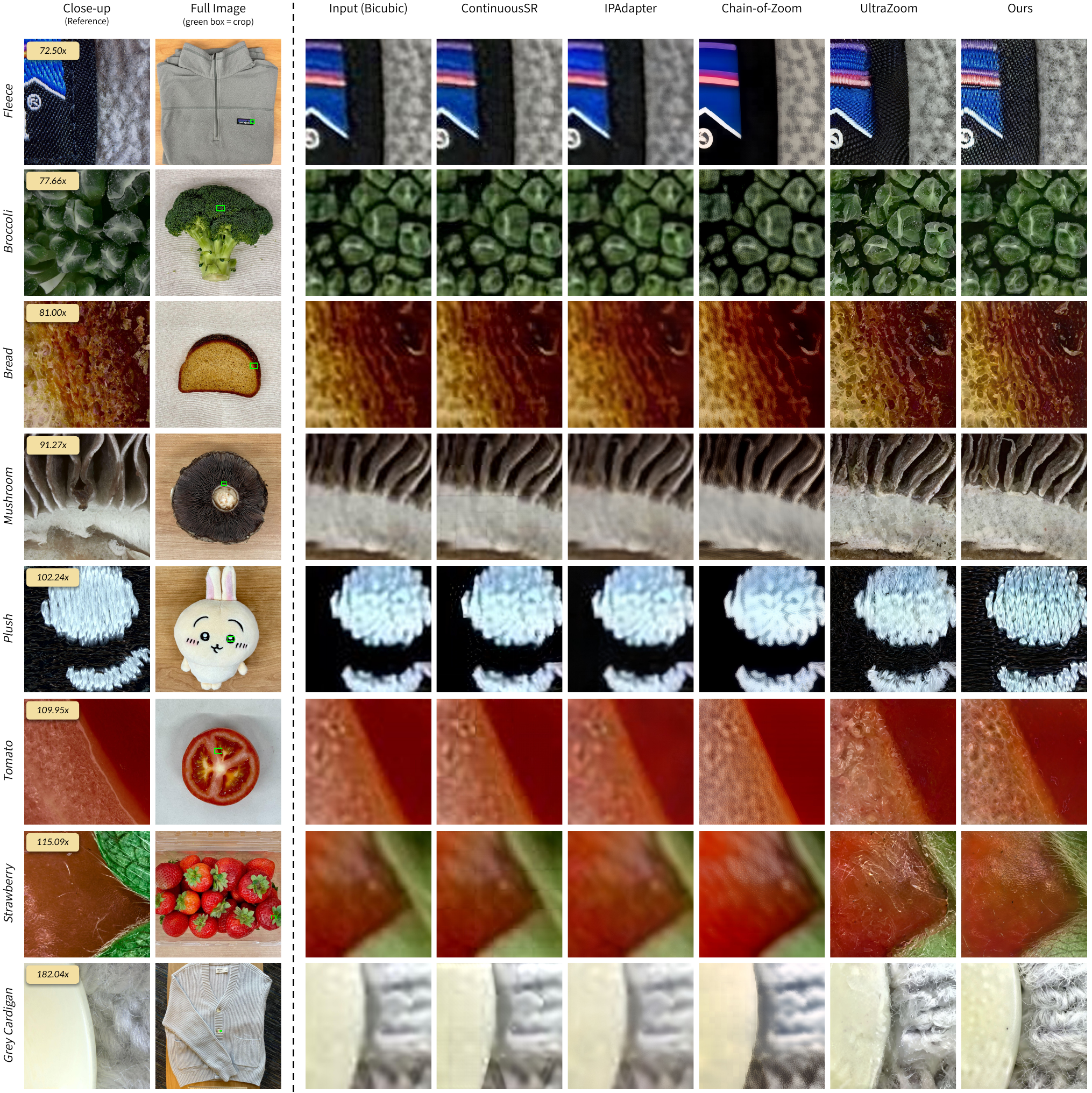}
    \caption{\textbf{Additional Multi-Material Examples.} We present 8 more examples of objects with multiple materials, demonstrating MicroZoom's ability to synthesize plausible details for scenes with varying textures and boundaries. For more results, see our \href{https://microzoom-sr.github.io/}{\textcolor{teal}{project page}}.}
    \label{fig:more_multi_texture}
\end{figure*}

\begin{figure*}[t]
    \centering
    \includegraphics[width=\textwidth]{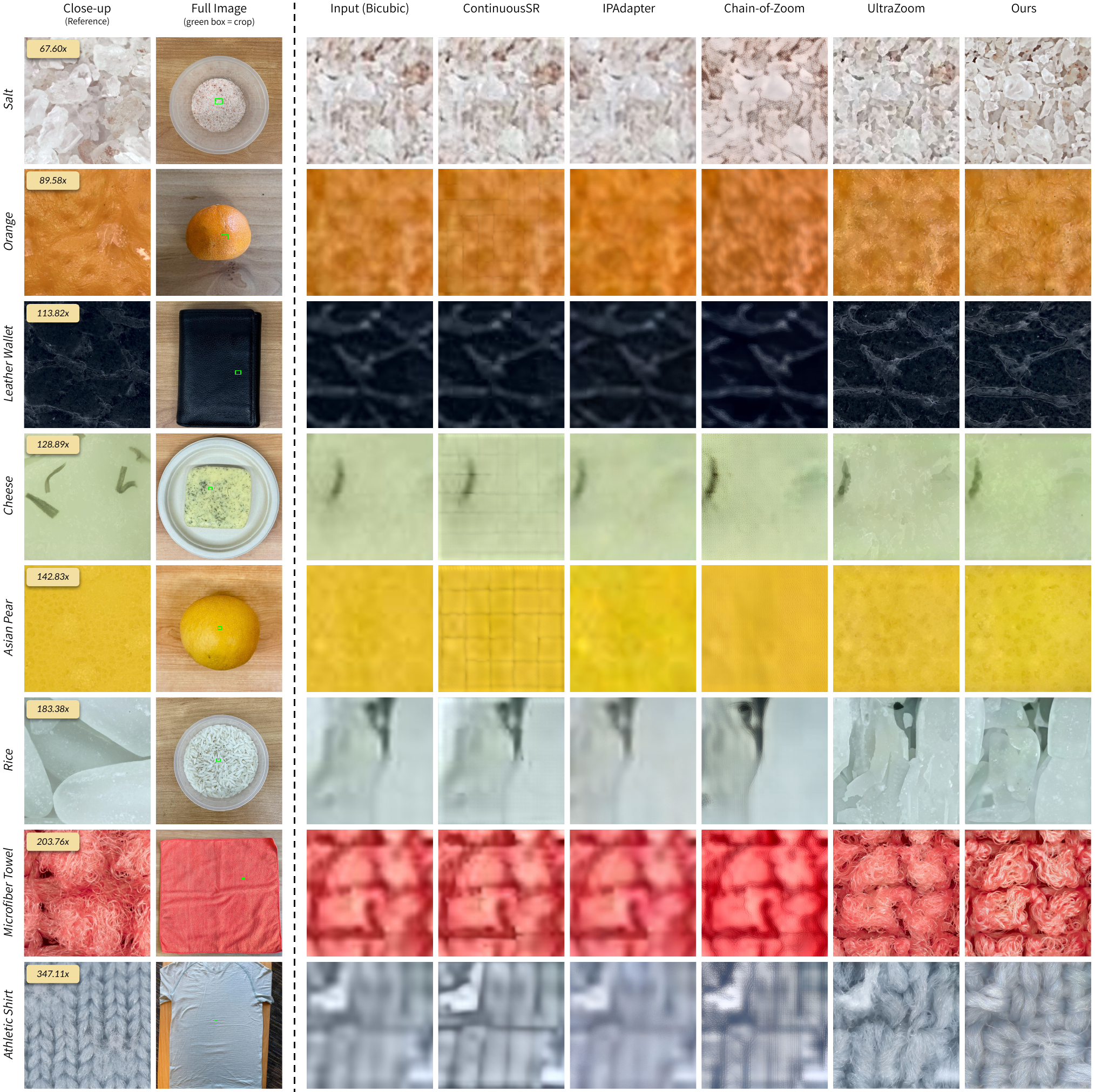}
    \caption{\textbf{Additional Single Material Examples.} We present 8 more examples of objects with a single material, demonstrating MicroZoom's ability to synthesize plausible textures for objects containing a single pattern type. For more results, see our \href{https://microzoom-sr.github.io/}{\textcolor{teal}{project page}}.}
    \label{fig:more_single_texture}
\end{figure*}

\appendix
\twocolumn[
    \noindent
    {\sffamily \huge Supplementary Material}
    \vspace{0.5cm}
]

\renewcommand{\thesection}{S-\arabic{section}}
\section{Log Spectral Distance}

To quantitatively evaluate the fidelity of high-frequency texture details synthesized by our method, we employ Log Spectral (LSD) metric. LSD measures the distance between the frequency spectra of the ground truth and the reconstructed images, directly capturing discrepancies in the frequency domain. It is defined as the root mean square of the differences between the logarithmic power spectra:
\begin{equation}
\mathrm{LSD}(\mathbf{x}, \hat{\mathbf{x}})
= \sqrt{\frac{1}{K} \sum_{k=1}^{K}
\left(
\log \bar{P}_{\mathbf{x}}(k)
-
\log \bar{P}_{\hat{\mathbf{x}}}(k)
\right)^2 } .
\end{equation}




\section{Consistent Synthesis}

\begin{figure}[h!]
    \centering
    \includegraphics[width=\linewidth]{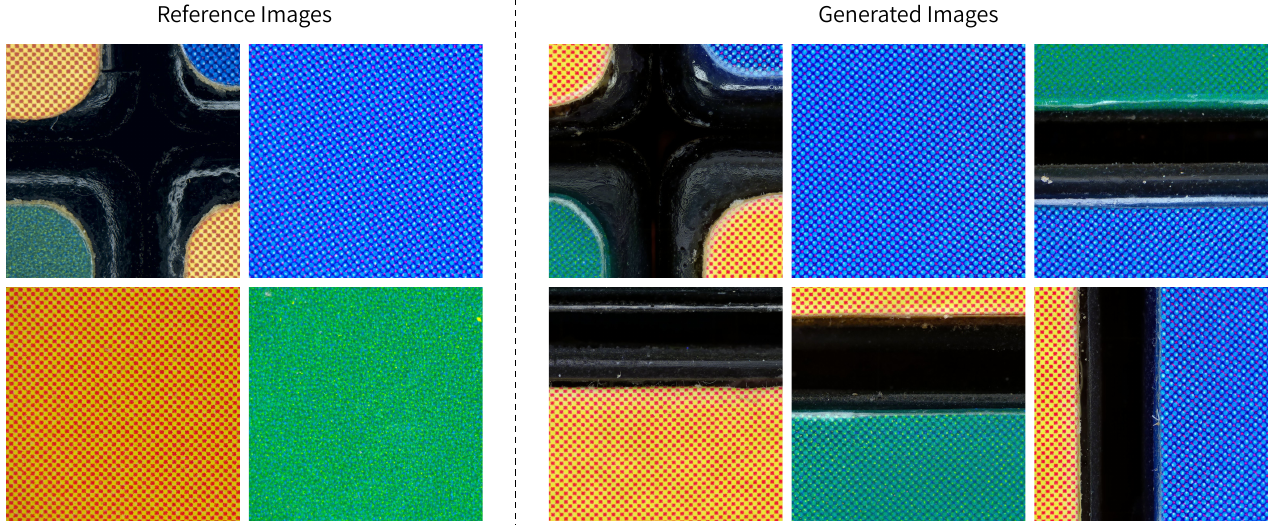}
    \caption{\textbf{Consistent Synthesis.} We present multiple distinct crops sampled from various locations within a single super-resolution output. Our method consistently generates plausible structures across the entire image, demonstrating robust performance across regions.}
    \label{fig:multi_crop}
\end{figure}

To evaluate the spatial reliability of our approach, we examine multiple distinct crops sampled from a single super-resolved output, shown in Figure \ref{fig:multi_crop}. Our method successfully maintains structural coherence and generates plausible, uniform high-frequency details across entirely different regions of the same image.

For more examples and continuous inspection of spatial consistency beyond isolated crops, interact with our  \href{https://duckyduck123.s3.amazonaws.com/index.html?ts=12345}{\textcolor{teal}{full-resolution demos}}.

\section{Synthetic LR Examples}

\begin{figure*}[t]
    \centering
    \includegraphics[width=0.93\textwidth]{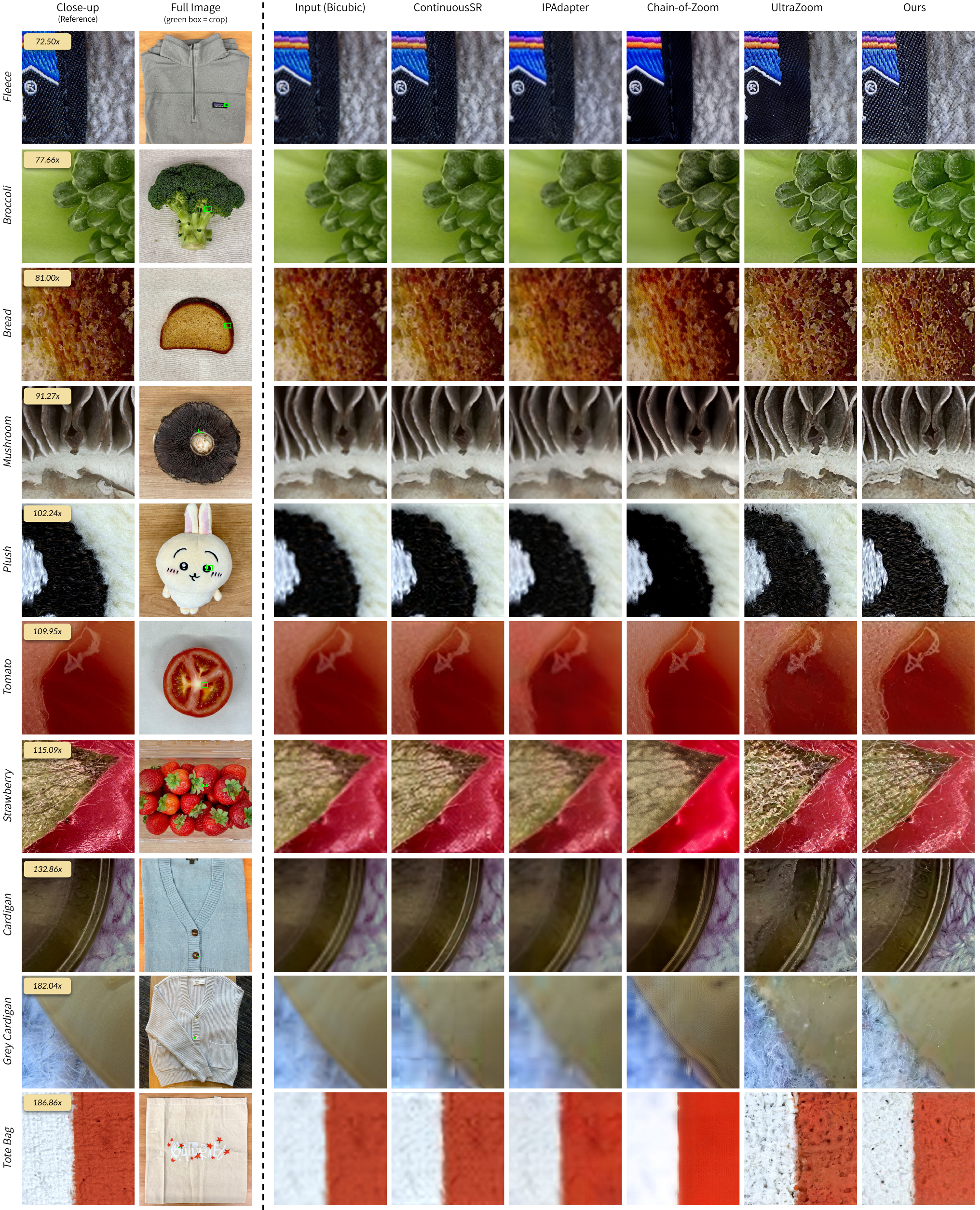}
    \caption{\textbf{Syn-LR Multi-Material Examples.} Full qualitative examples generated from synthetic LR inputs for objects containing multiple materials, corresponding to the quantitative evaluation in Table~\ref{tab:quant_table1}.}
    \label{fig:syn_lr_multi}
\end{figure*}

\begin{figure*}[t]
    \centering
    \includegraphics[width=0.93\textwidth]{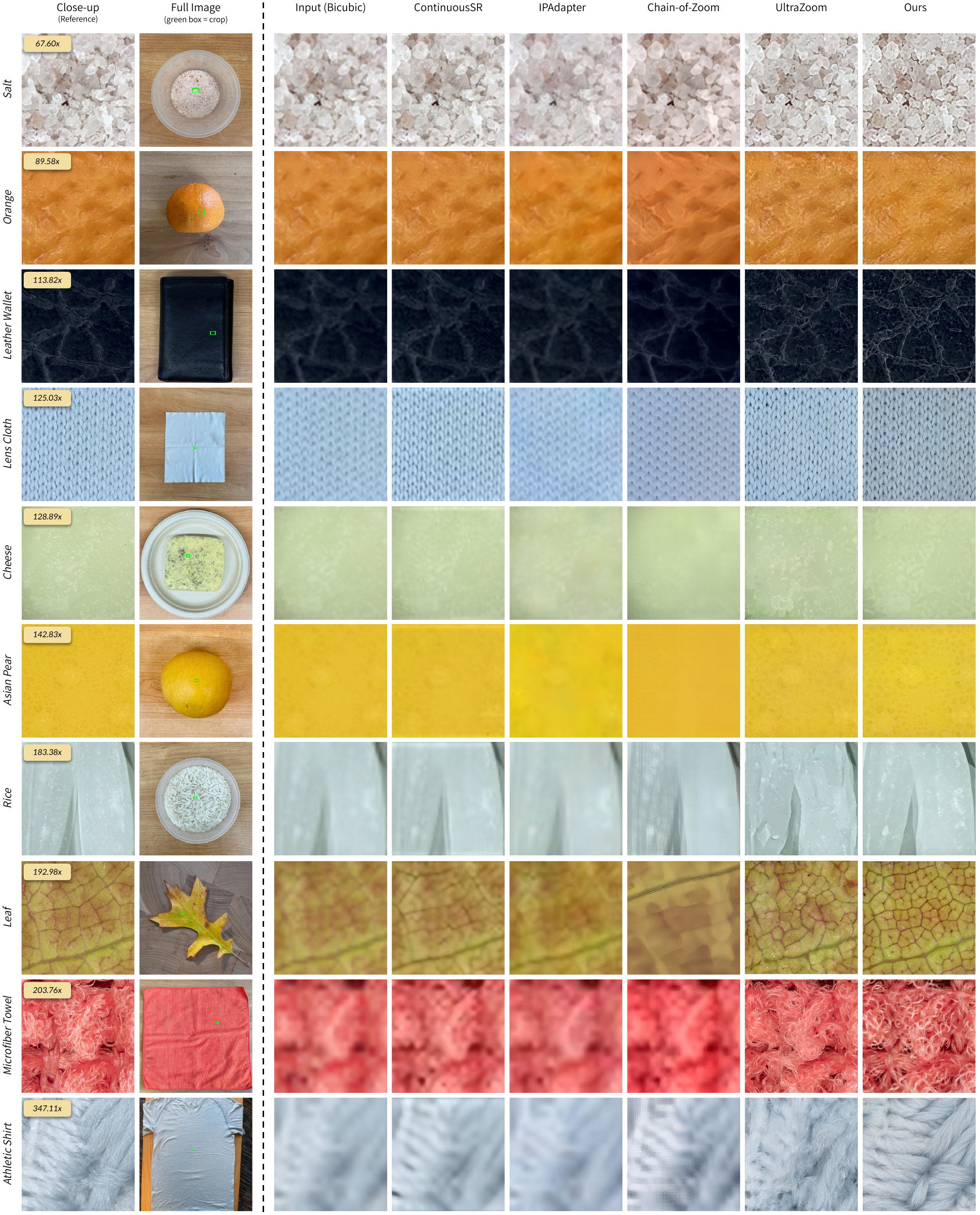}
    \caption{\textbf{Syn-LR Single Material Examples.} Full qualitative examples generated from synthetic LR inputs for objects containing a single material, corresponding to the quantitative evaluation in Table~\ref{tab:quant_table1}.}
    \label{fig:syn_lr_single}
\end{figure*}

In this section, we present the comprehensive set of generated results using synthetic low-resolution (Syn-LR) inputs, which we use for quantitative comparison in Table~\ref{tab:quant_table1}. Figure \ref{fig:syn_lr_multi} showcases all examples comprising of multiple materials, while Figure \ref{fig:syn_lr_single} provides the complete set of results for single-material objects.


\end{document}